\documentclass[11pt,a4paper]{article}
\usepackage[hyperref]{acl2020}
\usepackage{times}
\usepackage{latexsym}

\usepackage{booktabs}
\usepackage[demo]{graphicx}
\usepackage{caption}
\usepackage{subcaption}
\usepackage{microtype}
\usepackage{amsfonts}
\usepackage{upquote}
\usepackage{listings,lstautogobble}
\usepackage{color}
\usepackage{mathtools}
\usepackage{enumitem}
\usepackage{cprotect}
\usepackage{multirow}
\newif\ifshownlg
\shownlgtrue 

\definecolor{editorGray}{rgb}{0.95, 0.95, 0.95}
\definecolor{editorOcher}{rgb}{1, 0.5, 0} 
\definecolor{editorGreen}{rgb}{0, 0.5, 0} 
\definecolor{babyblueeyes}{rgb}{0.63, 0.79, 0.95}

\newcommand{\HTLM}{\texttt{HTLM}}

\newcommand\blfootnote[1]{%
  \begingroup
  \renewcommand\thefootnote{}\footnote{#1}%
  \addtocounter{footnote}{-1}%
  \endgroup
}

\lstdefinelanguage{JavaScript}{
  morekeywords={typeof, new, true, false, catch, function, return, null, catch, switch, var, if, in, while, do, else, case, break},
  morecomment=[s]{/*}{*/},
  morecomment=[l]//,
  morestring=[b]"
}
\lstdefinelanguage{HTML5}{
        language=html,
        sensitive=true, 
        alsoletter={<>=-},
        otherkeywords={
        <html>, <head>, <title>, </title>, <meta, />, </head>, <body>, <html, <div, lang, xml:lang,
        <canvas, \/canvas>, <script>, </script>, </body>, </html>, <!, html>, <style>, </style>, ><
        },  
        ndkeywords={
        =,
        charset=, id=, width=, height=,
        border:, transform:, -moz-transform:, transition-duration:, transition-property:, transition-timing-function:, <mask>, <mask>12, <mask>X
        },  
        morecomment=[s]{<!--}{-->},
        tag=[s],
        autogobble=true
}

\aclfinalcopy 

\title{HTLM: Hyper-Text Pre-Training and Prompting of Language Models}

\newcommand{\affilsup}[1]{\rlap{\textsuperscript{\normalfont#1}}}
\author{Armen Aghajanyan\affilsup{1},~\textsuperscript{$*$} ~~Dmytro Okhonko\affilsup{1},~\textsuperscript{$*$} 
~~Mike Lewis\affilsup{1}, ~~Mandar Joshi\affilsup{1,2}, \\
\textbf{Hu Xu\affilsup{1}, ~~Gargi Ghosh\affilsup{1}, ~~Luke Zettlemoyer\affilsup{1,2}} \\
\textsuperscript{1}Facebook AI ~~~~
\textsuperscript{2}University of Washington \\
  \texttt{\{armenag,oxo,mikelewis,mandarj,huxu,gghosh,lsz\}@fb.com} 
  }

\date{}
\begin{document}
\maketitle
\begin{abstract}
\blfootnote{\textsuperscript{$*$}~Equal Contribution}

We introduce \HTLM{}, a hyper-text language model trained on a large-scale web crawl. 
Modeling hyper-text has a number of advantages: (1) it is easily gathered at scale, (2) it provides rich document-level and end-task-adjacent supervision (e.g. \verb+class+ and \verb+id+ attributes often encode document category information), and (3) it allows for new structured prompting that follows the established semantics of HTML (e.g. to do zero-shot summarization by infilling \verb+<title>+ tags for a webpage that contains the input text). 
We show that pretraining with a BART-style denoising loss directly on simplified HTML provides highly effective transfer for a wide range of end tasks and supervision levels. 
\HTLM{} matches or exceeds the performance of comparably sized text-only LMs for zero-shot prompting and fine-tuning for classification benchmarks, while also setting new state-of-the-art performance levels for zero-shot summarization. 
We also find that hyper-text prompts provide more value to \HTLM{}, in terms of data efficiency, than plain text prompts do for existing LMs, and that \HTLM{} is highly effective at auto-prompting itself, by simply generating the most likely hyper-text formatting for any available training data.
We will release all code and models to support future \HTLM{} research. 
\end{abstract}

\section{Introduction}
\begin{figure}
\centering
\begin{subfigure}[b]{0.45\textwidth}
   \centering
    \begin{lstlisting}[numbers=none, basicstyle=\tiny\ttfamily]
        <!DOCTYPE html>
        <html>
            <title> <mask>12 </title>
            <body>
               ~ south korea on monday announced sweeping tax reforms , including income and corporate tax cuts to boost growth by stimulating sluggish private consumption and business investment .
            </body>
        </html>
    \end{lstlisting}
    $\downarrow$
\end{subfigure}
\begin{subfigure}[b]{0.45\textwidth}
   \centering
   \begin{lstlisting}[numbers=none, basicstyle=\tiny\ttfamily]
        <!DOCTYPE html>
        <html>
            <title> ~ South Korea Announces Tax Reforms To Boost Economic Growth ~ </title>
            <body>
               ~ south korea on monday announced sweeping tax reforms...
            </body>
        </html>
    \end{lstlisting}
\end{subfigure}

\caption{An example structured prompt for a simple summarization task, where we ask a generative masked language model to generate a mask representing the title with an average tokens size of 12.}
\label{fig:example_prompt}
\end{figure}

The vast majority of text used to pretrain language models is extracted from web pages, while discarding any markup they contain~\cite{ROBERTA,gpt3,T5,BART}. We argue that this HTML should not be ignored; it enables new forms of highly effective language model pretraining and prompting with structured document-level supervision.

Hyper-text, such as the HTML found in the Common Crawl\footnote{\url{https://commoncrawl.org/}}, has a number of advantages for pretraining over plain text. It often encodes high-level properties of different parts of the documents, which are difficult to infer from the text alone. For example, \verb+<title>+ elements can be excellent summaries of the \verb+<body>+ of a document, while element \verb+class+ and \verb+id+ attributes can encode categorical properties of  documents. Such supervision is highly diverse, depending on what the website authors choose to present, and provides close proxies for many NLP tasks we aim to later solve. 

Modeling hyper-text allows us to introduce \emph{structured prompting} of language models. We design prompts that incorporate the established semantics of HTML to better control for the desired model output. This includes, for example, performing zero-shot summarization by asking the model to infill \verb+<title>+ tags in a web page. And, the fact that we jointly model text and hyper-text formatting also allows for effective auto-prompting. If we have even a few examples for a new task, we can directly ask the model to format them in HTML, and templatize the result to define the new prompt. 

Our \textbf{H}yper\textbf{T}ext \textbf{L}anguage \textbf{M}odel (\HTLM{}) is trained on 23TB of simplified HTML which we automatically extract from common crawl dumps (see Section~\S\ref{sec:mhtml}). We use a modified BART denoising objective~\cite{BART} that randomly masks spans of hyper-text and aims to reconstruct the original input. We extend the original masking with a new size hint scheme, where each mask is associated with an integer that provides a noisy hint for the size of the masked text, to allow for more fine grained task-specific length priors when prompting the final model (see Section~\S\ref{sec:model:size-hints}). Figure~\ref{fig:example_prompt} shows an example mask that should be reconstructed with a phrase that contains roughly 12 tokens. 

Through extensive experiments, we show that our \HTLM{} achieves highly effective transfer for a wide range of end tasks and supervision levels. 
It matches or exceeds the performance of comparably sized text-only LMs for zero-shot prompting and full fine-tuning on GLUE, while also setting new state-of-the-art performance levels for zero-shot summarization with a gain of up to 8 ROUGE-1 points. 
It also allows few shot learning for problems that are less easily reduced to text-only inputs, such table to text generation. Following methodology introduced by \citet{how_many_datapoints}, we further find that hyper-text prompts provide more data efficiency to the \HTLM{} model than plain text prompts do for existing LMs, being effectively equivalent to having up to a thousand extra training examples. Finally, we see that the \HTLM{} model is highly effective at auto-prompting itself, in some cases rivaling the performance of manually engineered prompts. 

In summary, our contributions include:
\begin{itemize}
    \item We present the first hyper-text language model (\HTLM{}), trained on 23TB of simplified HTML data from the common crawl. 
    \item Our new hyper-text prompting scheme uses both the well-established semantics of HTML and new size hints on prompt masks to provide more fine-grained control of new task specifications. 
    \item We demonstrate consistently strong transfer from \HTLM{} to a range of tasks at differing supervision levels, including improving the best-known zero-shot summarization numbers by up to 8 ROUGE-1 points.
    \item Following \citet{how_many_datapoints}, our data efficiency analysis shows that hyper-text prompts are worth more to the \HTLM{} model than plain text prompts are for existing LMs, being effectively equivalent to having up to a thousand extra training examples. 
    \item We demonstrate the \HTLM{} directly supports auto prompting for new tasks, by simply asking it to format any available examples in HTML, often rivaling or surpassing previous manually engineered prompts. 
    \item We release all code and models to support future \HTLM{} research.
\end{itemize}

\section{HyperText Language Model (\HTLM{})}
\HTLM{} is trained on a large corpus of simplified HTML, which is automatically extracted from the common crawl (Section~\S\ref{sec:mhtml}). We use a BART-style denoising autoencoder with span masking (Section~\S\ref{sec:model}), extended to allow size hints during reconstruction of the original text (Section~\S\ref{sec:model:size-hints}). 

\subsection{Minimal HTML}
\label{sec:mhtml}
Although HTML contains supervision signals to natural language, the majority of HTML in a modern web page does not provide any significant form of supervision for pretraining. For example, a large portion of a webpage is JavaScript code or CSS, which provides more aesthetics to the page rather than document-level information. Coupling this with the challenges of training transformers on very long sequence lengths \citep{performer, linformer, longformer}, it was important to automatically convert web pages to a simplified form, which we call \textbf{M}inimal-\textbf{HTML} (MHTML), as defined below.

We remove all sub-trees of the HTML DOM\footnote{The DOM or Document Object Model is an interface that treats an HTML document as a tree structure wherein each node is an object representing a part of the document.} which do not contain textual elements of a certain character size (128 for standard textual elements, 64 for lists/tables/spans). 
We also filter out all \textit{headers}, \textit{footers}, \textit{copyrights}, \textit{forms}, and \textit{iFrames}. We fold consecutive \verb+<div>+ elements into a singular \verb+<div>+ element with merged attributes. We also remove all attributes which are not \verb+class+ or \verb+id+ attributes. Lastly, we skip all MHTML documents whose ratio of text to HTML is not greater than $0.46$. Particularly we noticed that MHTML documents whose ratio of text to HTML is low, the average quality of the document tends to be lower as well. We found these numbers by visually inspecting a set of Common Crawl (CC) documents after application of aforementioned transforms ensuring both a high quality of kept documents while also not filtering too large amount of data. Furthermore we filter out all documents who have a \verb+lang+ attribute that is not set to \verb+en+.

Applying these deterministic transformations removes on average 94\% of characters from a raw webpage while maintaining the general markup of the document. Furthermore, it allowed close to 85\% of MHTML documents to fit into 1024 BPE tokens; the maximum token length for BART and many other existing language models.

One by-product of this type of filtering is that it also produced high-quality documents by default\footnote{Much of the noise in existing text collections derived from the common crawl comes from artifacts that are introduced when returning the text in the relatively arbitrary order it appeared in the original HTML, before the markup was stripped.}; thus, we opted out of model-based filtering of documents such as CC-100 \citep{XLMR}. We used the January 2021 snapshot of Common Crawl, which provided us with 23 Terabytes of MHTML text after filtering.

\subsection{Model}
\label{sec:model}

We adopt a BART-style denoising auto-encoder~\citep{BART} for several reasons. We want to predict arbitrary substrings within the MHTML, conditioned on the rest of the document. This allows us to equally easily (1) use masks during prompting to mark where to generate text associated with model outputs within a web page, and (2) automatically generate prompts by wrapping plain text training examples in masks that allow the model to mark them up by generating MHTML formatting.  We also do not know in advance exactly how much text needs to be generated in each case, thereby ruling out the use of more traditional masked language models. 

For all of our experiments, we adopt the same architecture as BART-Large and initialized our models with the BART-Large checkpoint. This model has roughly 400 million parameters.

We trained our augmented BART model for a total of 330,000 steps on 256 GPUs with an effective batch size of 8192. We initialize our model with the original BART-Large model. We train using the Adam optimizer~\citep{ADAM} and a polynomial decay learning rate scheduler with a peak learning rate of $4\mathrm{e}{-5}$ and $10,000$ warm-up steps.

We do not use the sentence shuffling from the original BART objective, and select a Poisson $\lambda$ of 3.5 for sampling span lengths for masking. We set dropout in the attention to $0.1$ for the first 170k steps, reducing it to $0.0$ thereafter. We also filter out data to only English (\verb+en+) after 170k steps using FastText \cite{fasttext}. We noticed the perplexity plateaued around 170k steps which is why we simplify the learning process by removing dropout and applying stronger filtering of the English language. 

\subsection{Size Hints}
\label{sec:model:size-hints}

BART allows each mask to be replaced with multiple tokens during the reconstruction. During pre-training, BART masks a span with the length sampled from a Poisson distribution; thus, the model must learn to implicitly predict the length of the masked text. A fundamental problem we encountered when trying to use standard BART for zero-shot generative prompting is the inability to control the length of the generated text for each mask, even when using various decoding strategies like length penalties. 

To allow for more control, we augment BART's masking scheme by introducing size hints. Specifically, we tokenize the noisy estimate of the length of a span directly and insert it right after the span mask token. For example, given the correct mask length $m$, we insert $n$ \textit{$\left<mask\right>$} tokens where $n$ is $\max\left(1,\lfloor\mathcal{N}(m, m * \epsilon)\rfloor\right)$ and $\epsilon$ is a hyperparameter representing how noisy we want these size hints to be. By optionally injecting size hints, we can prompt the model to generate text of roughly some specific length, or by not injecting size hints, we allow the model to model the mask size implicitly. We give size-hints to $80\%$ of masks with the noisiness of size hints $\epsilon=0.1$. 

We provide an example of the benefits of size hints in generation in Table~\ref{table:size_hint_sample}.

\begin{table*}[!htpb]
\centering
\begin{tabular}{p{8.3cm}p{2.0cm}p{4.6cm}}
\toprule
Prompt & Size Hint (X) & \HTLM{} Output \\ \midrule
        \begin{lstlisting}[numbers=none, basicstyle=\tiny\ttfamily, breaklines=true]
    <html id="cnn_dailymail" lang="en" xml:lang="en">
        <head class="pg-headline" data-act-id="article_head_0">
            <p> <mask>X --CNN</p>
        </head>
        <body>
            <p>
            However, observers inside the court said the prosecution evidence was difficult to follow or to understand. Correspondents said the most prominent video clip was by another al-Jazeera journalist who is not on trial. The three, including ex-BBC reporter Peter Greste, deny spreading false news and helping the Muslim Brotherhood. They appeared in the Cairo court on Tuesday along with other Egyptian students and activists being tried with them. Defence lawyers complained they were unable to understand some of the recordings being played because of poor technical quality. While the recordings were played, defendants talked among themselves in the caged dock. The trial was later adjourned until 3 May and bail for the defendants was denied, reports from the court said. Peter Greste, producer Baher Mohamed and al-Jazeera English's Canadian-Egyptian bureau chief Mohamed Fadel Fahmy have been held since December. A fourth al-Jazeera reporter, Abdullah Elshamy, who works for the network's Arabic channel, has been detained since August but not charged. Egyptian authorities accuse the journalists of aiding the Islamist Muslim Brotherhood, which has been banned as a terrorist group.
            </p>
        </body>
    </html>
    \end{lstlisting}  &     \vspace{1em} 5  & \vspace{1em} Three Egyptian journalists are being tried --CNN                         \\
        &       \vspace{-20em} 8                    & \vspace{-20em} Three British journalists accused of aiding the Muslim Brotherhood are tried --CNN      \\
       &         \vspace{-16em} 12                     & \vspace{-16em} Three British journalists accused of aiding the Muslim Brotherhood are on trial --CNN  \\ 
       & \vspace{-11em} 15 & \vspace{-11em} Three British-Egyptian journalists accused of aiding the Muslim Brotherhood are to be tried in a Cairo court --CNN \\ [-5em] \bottomrule
\end{tabular}
\caption{We provide a simple example using our CNN/DM prompt where by altering the Size Hint value (X) we get summaries of varied length and complexity. }
\label{table:size_hint_sample}
\end{table*}

\section{HTML-based Prompting}
We use the HTML-based prompting scheme for a range of generation and classification tasks. Broadly, we use HTML templates--either selected manually or generated by the model itself by auto-prompting--to specify the HTML structure of the task. The template is then instantiated with the task input and placeholder mask tokens for the output. The model uses this instantiated template as a prompt.
Because BART models reconstruct the full input, we rely on simple heuristics to match the prefix/suffix around any masks and extract the final output.

\subsection{Generation Prompting Policies}
Given that we have optional size hints for masks, a single prompt can generate a wide variety of text; therefore, we discuss multiple policies to select the prompted results. We can decide not to utilize size hints at all and thus remove the need to use any policies, but this comes at the cost of template robustness. Without size hints, a template not only has to express the semantics of the task, but also needs to match the average target length as well; such prompts are brittle and require careful manual design. However, using hints allows us to decouple generation length from the prompt, greatly improving template reuse across related tasks. It is also possible that for a prompt and a specific subset of the data, \HTLM{} will not generate an output from which we can programmatically extract the generated mask; therefore, our policies for size-hints also mitigate this issue.

For every generation task, we first construct a prompt that can generate the correct text semantically, and then we provide size hints equal to the average target of a subset of the training set, $\bar s$. If, for a particular input, we are not able to extract a value, we run \HTLM{} on the same prompt, but with our size hint set to $\bar s  \pm i\epsilon \bar s$, from which we select the output with the lowest perplexity, we continue this process at most five times where $i$ represents the current index of the policy. If we still cannot find a valid generated answer, we fall back on the auto-template described in the next section. In experiments, we denote \HTLM{}-Manual-NS (not sized) as our manually engineered prompt with no size hint, while \HTLM{}-Manual-S uses the policy defined here for all generation benchmarks.

\subsection{Auto-Prompting}
To avoid manually engineering prompts, we also explore automatic generation of structured prompts.
By training on hypertext, \HTLM{} can learn high-level document semantics that we exploit for prompt creation. We generate prompting templates by asking the model to recover document markups. Specifically, we place \textit{$\left<mask\right>$} tokens around every independent block of data (e.g. summary/article).

\begin{figure*}[!htbp]
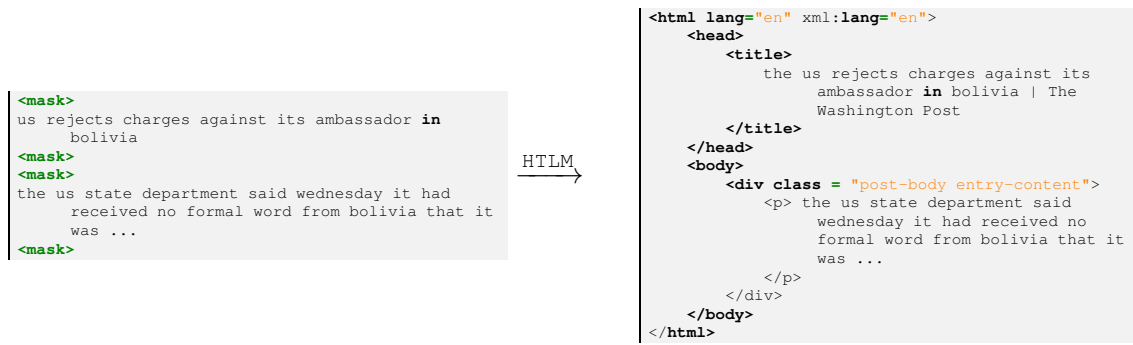

\centering

\begin{subfigure}{.45\textwidth}
    \centering
    \begin{lstlisting}[numbers=none, basicstyle=\tiny\ttfamily]
        <mask> 
        us rejects charges against its ambassador in bolivia 
        <mask>
        <mask> 
        the us state department said wednesday it had received no formal word from bolivia that it was ... 
        <mask>
s    \end{lstlisting}
\end{subfigure} $\xrightarrow{\text{\HTLM{}}}$
\begin{subfigure}{.45\textwidth}
    \centering
    \begin{lstlisting}[numbers=none, basicstyle=\tiny\ttfamily]
        <html lang="en" xml:lang="en">
            <head>
                <title>
                    the us rejects charges against its ambassador in bolivia | The Washington Post
                </title>
            </head>
            <body>
                <div class = "post-body entry-content"> 
                    <p> the us state department said wednesday it had received no formal word from bolivia that it was ...
                    </p>
                </div>
            </body>
        </html>
    \end{lstlisting}
\end{subfigure}
\cprotect\caption{An example of auto-prompting using a sample from the train-set of the Gigaword dataset. \HTLM{} places the summary inside of a \verb+<title>+ inside of a \verb+<head>+ element, while placing the article in a \verb+<div>+ element with an \verb+entry-content+ attribute value for attribute \verb+class+ which is common on news web-sites.}
\label{fig:auto_prompting}
\end{figure*}

We provide an example of auto-prompting for a sample from the Gigaword summarization dataset~\citep{gigaword} with the respective masking in Figure~\ref{fig:auto_prompting} . For our generation experiments, we denote \HTLM{}-Auto-NS (not-sized) as the auto-prompt without using size hints, where \HTLM{}-Auto-S uses the size hints based policy described in the previous section.

We found that \HTLM{} auto-prompting was less effective for classification tasks. We hypothesize that this is because generative targets carry significantly more information than a simple binary target token.

\section{Zero/One-Shot Prompting}
\citet{true_few_shot_learning} argue that zero/few-shot learning cannot happen when prompts are created by tuning on a large amount of development data. To mitigate for this issue all the manual prompts used throughout our experiments are either derived from related papers or developed using a maximum of fifty samples from the train set.

\subsection{Generation}
We evaluate \HTLM{} on summarization, a prototypical generation task. For all summarization benchmarks, we use ROUGE-1/2/L as our primary metrics to stay consistent with other literature~\citep{ROUGE}. 

\ifshownlg
Furthermore we benchmark \HTLM{} on a set of three standard natural language generation tasks. We utilize the official benchmarking scripts provided which report BLEU \citep{bleu}, NIST \citep{NIST}, METEOR \citep{METEOR}, ROUGE-L \citep{ROUGE},
CIDEr \citep{cider} and TER \citep{ter}. We use \citet{prefixtuning} for our baselines, and present prefix tuning results with 0.1\% of parameters as well.
\fi
\begin{itemize}[label={},leftmargin=0pt]
\item \textbf{Gigaword} consists of headlines from news articles~\citep{gigaword}. The target summaries are relatively short, consisting roughly on average of 10 BPE tokens.

\item \textbf{CNN/Dailymail}~\citep{cnndailymail}
 provides multi-sentence target summaries close to 3 sentences, or roughly 50 tokens.

\item \textbf{Reddit TIFU}~\citep{reddittifu}
contains summaries of Reddit posts. Specifically, we use the \textit{short} subset of data . Compared to our other summarization datasets, this dataset is highly abstractive and not based on news articles.

\item \textbf{XSum} \citep{xsum} provides abstractive single sentence summaries of news articles.

\ifshownlg
\item \textbf{E2E} \cite{e2e_nlg} is a table-to-text generation dataset containing approximately 50K
samples with 8 unique fields from the restaurants domain. 
\item \textbf{WebNLG} \cite{webnlg} is also a structured generation dataset containing 15 different domains from DBPedia. We report numbers on the Seen (S), Unseen (U) and All (A) subsets of the data.
\item \textbf{DART} \cite{dartnlg} is a open-domain structured generation dataset containing Wikipedia tables.
\fi
\end{itemize}
We manually searched for prompts for each of these datasets using a maximum of 50 data points from the train set to evaluate the prompts.
For our baseline, we compare against PEGASUS~\citep{pegasus}, the current state of the art for zero shot summarization. PEGASUS was explicitly pre-trained for summarization by masking and generating salient \emph{gap} sentences from news articles.
We present our results in Table~\ref{table:HTLM_summarization}.

\begin{table*}[htpb]
\centering
\begin{tabular}{@{}lcccc@{}}
\toprule
Model        & Gigaword          & CNN/DM            & Reddit TIFU      & XSum             \\ \midrule
PEGASUS-0S   & 23.39/07.59/20.20  & 32.90/13.28/29.38 & 14.66/3.06/10.17 & 19.27/3.00/12.72 \\ \midrule
\HTLM{}-Auto-NS & 27.56/10.17/24.57 & 33.40/13.45/30.10 & 6.71/1.98/7.86   & 15.15/2.54/10.91 \\
\HTLM{}-Auto-S  & 28.73/11.31/26.49 & 34.65/14.54/32.15 & 8.15/2.92/9.75   & 17.14/3.41/13.43 \\
\HTLM{}-Manual  & \bf{31.61/10.80/28.60} & \bf{38.51/16.10/33.89} & \bf{15.81/2.98/10.54} &             \bf{22.34/4.12/14.56}     \\ \bottomrule
\end{tabular}
\caption{\HTLM{} results on zero-shot summarization. \HTLM{}-Manual denotes manually engineered prompts with size hints, while \HTLM{}-Auto-S and \HTLM{}-Auto-NS indicate autoprompting with and without size hints respectively. Metrics shown are ROUGE-1/ROUGE-2/ROUGE-L respectively.}
\label{table:HTLM_summarization}
\end{table*}

\ifshownlg
\begin{table*}[h]
    \centering
    \resizebox{2.1\columnwidth}{!}{
    \addtolength{\tabcolsep}{-2pt} 
    \begin{tabular}{lccccc|ccccccccc|cccccc}
\toprule 
& \multicolumn{5}{c}{E2E} & \multicolumn{9}{|c}{WebNLG} & \multicolumn{6}{|c}{DART} \\
         & BLEU  & NIST & MET & R-L & CIDEr  & \multicolumn{3}{c}{BLEU}  & \multicolumn{3}{c}{MET} & \multicolumn{3}{c|}{TER $\downarrow$}  & BLEU   & MET   & TER $\downarrow$  & Mover    & BERT   & BLEURT \\
& \multicolumn{5}{c|}{ } & S & U & A & S & U & A & S & U & A \\
\midrule
\midrule
Fine-tuning & \multicolumn{20}{c}{} \\
\quad \quad GPT-2$_{\textsc{MEDIUM}}$       & 68.2 & 8.62 & 46.2  & \textbf{71.0}  & 2.47 & 
64.2  & 27.7 & 46.5 & 0.45 & 0.30 & 0.38 & \textbf{0.33} & 0.76 & 0.53 & 
46.2 & \textbf{0.39} & 0.46 & \textbf{0.50} & \textbf{0.94} & 0.39 \\ 
\quad \quad GPT-2$_{\textsc{LARGE}}$       & 68.5 & 8.78 & 46.0 & 69.9 & 2.45 & 65.3 & 43.1 & 55.5 & \textbf{0.46} & 0.38 & \textbf{0.42} & \textbf{0.33} & 0.53 & 0.42  
& 47.0 & \textbf{0.39} & 0.46 & \textbf{0.51} & \textbf{0.94} & \textbf{0.40} \\ 
\quad \quad \HTLM{}       & \textbf{70.3} & \textbf{8.90} & \textbf{46.3} & 70.8 & \textbf{2.47} & \textbf{65.4} & \textbf{48.4} & \textbf{55.6} & \textbf{0.46} & \textbf{0.39} & \textbf{0.42} & \textbf{0.33} & \textbf{0.51} & \textbf{0.40}
& \textbf{47.2} & \textbf{0.39} & \textbf{0.44} & \textbf{0.51} & \textbf{0.94} & \textbf{0.40} \\ 
\midrule
Prefix (0.1\%) & \multicolumn{20}{c}{} \\
\quad \quad GPT-2$_{\textsc{MEDIUM}}$ & 69.7 & 8.81 & 46.1  & 71.4  & 2.49 & 62.9 & 45.6 & 55.1 & 0.44 & 0.38 & 0.41 & 0.35 & 0.49 & 0.41 & 
46.4 & 0.38 & 0.46 & \textbf{0.50} & \textbf{0.94} & 0.39 \\ 
\quad \quad GPT-2$_{\textsc{LARGE}}$           & \textbf{70.3}  & \textbf{8.85}  & \textbf{46.2}  & \textbf{71.7}  & \textbf{2.47} & 63.4 & 47.7 & \textbf{56.3} & 0.45 & \textbf{0.39} & \textbf{0.42} & 0.34 & 0.48 & 0.40
& 46.7 & \textbf{0.39} & \textbf{0.45} & \textbf{0.51} & \textbf{0.94} & \textbf{0.40} \\
\quad \quad \HTLM{}        & 70.1 & \textbf{8.85} & 46.1 & 71.2 & 2.45 & \textbf{64.8} & 46.1 & \textbf{56.3} & \textbf{0.46} & 0.38 & \textbf{0.42} & \textbf{0.33} & \textbf{0.47} & \textbf{0.40}
& \textbf{47.1} & \textbf{0.39} & \textbf{0.45} & 0.50 & \textbf{0.94} & 0.39 \\ \midrule
One-Shot & \multicolumn{20}{c}{} \\
\quad \quad \HTLM{}       & 32.1 & 3.35 & 24.1 & 31.6 & 0.78 & 28.1 & 18.5 & 22.8 & 0.24 & 0.21 & 0.12 & 0.78 & 0.79 & 0.78 & 22.1 & 0.12 & 0.91 & 0.25 & 0.78 & 0.22 \\ \midrule
Base-lines & \multicolumn{20}{c}{} \\
\quad \quad TILB-Pipeline      & - & - & - & - & - & 44.34 & 20.65 & 35.29 & 0.38 & 0.21 & 0.30 & 0.48 & 0.64 & 0.56 & - & - & - & - & - & - \\ 
\quad \quad UIT-VNU-Pipeline & - & - & - & - & - & 19.87 & 0.11 & 7.07 & 0.15 & 0.03 & 0.09 & 0.78 & 0.87 & 0.82 & - & - & - & - & - & - \\ 

\bottomrule
\end{tabular}
}
\caption{We evaluate GPT-2$_{\textsc{MEDIUM}}$, GPT-2$_{\textsc{LARGE}}$ and \HTLM{} on table-to-text generation on E2E (left), WebNLG (middle) and DART (right). }
\label{tab:table-to-text} 
\end{table*}
\fi

\HTLM{} with manual prompts (\HTLM{}-Manual) and size hints substantially improves over state-of-the-art zero-shot summarization results on all four datasets without any tailored pretraining. In particular, we see large improvements of more than 8 ROUGE-L F1 for the Gigaword dataset. Furthermore, size hints-based auto-prompting (\HTLM{}-Auto-S) outperforms PEGASUS in three out of four datasets. Specifically, for the Gigaword dataset, we outperform previous state-of-the-art zero-shot results from PEGASUS by roughly 6 ROUGE-L points. 
\HTLM{} improvements stem from the fact that HTML-based prompting allows us better control over dataset-specific attributes such as length and style.

\ifshownlg
For NLG tasks, we required the use of a single training example to get prompting to work sufficiently. We report these one-shot numbers in Table~\ref{tab:table-to-text}. Because these tasks require structured tabular inputs, it is not obvious how to prompt any other text-based pre-trained models. We report other non-trainable baselines such as the grammar based pipeline approaches (TILB/UIT-VNU) in \citet{webnlg}. To the best of our knowledge, these are the first one-shot table to text, natural language generation results. 
\fi
\subsection{Classification}

For prompting in the classification setting, we select 4 datasets to work with. Instead of relying on generative prompting to generate target token(s) denoting the correct class, we instead rely on perplexity measures over the set of all targets to select the correct class. In other words, we select the class for which the perplexity of the corresponding instantiated template is the  smallest.

\begin{itemize}[label={},leftmargin=0pt]
\item \textbf{RTE}~\citep{rte} is a textual entailment task formulated as binary classification. We place the candidate in a \verb+<div>+ element with the class attribute set to \textit{candidate} and do the same with the respective hypothesis. In the third element, we utilize the prompt from \citet{gpt3} with the class attribute set to \textit{answer}.

\item \textbf{BoolQ}~\citep{clark2019boolq} is a yes/no question answering task, also formulated as binary classification for question, passage, and answer triplets. We represent the question as a \verb+<div>+ element with the itemprop set to \textit{https://schema.org/Question}, passage as a \emph{div} element with class attribute \textit{passage} and answer as a \emph{div} element with the itemprop set to \textit{https://schema.org/Answer}.

\item \textbf{Winogrande}~\citep{winograd} consists of adversarially collected Winograd Schema Challenge~\cite{levesque2011winograd} data. We utilize the same template as GPT-3 but place it in a QA style template similar to BoolQ. Please refer to the Appendix for exact templates.

\item \textbf{HellaSwag}
The last dataset we evaluate is the commonsense natural language inference task HellaSwag which, due to its adversarial nature, is considered complex \citep{hellaswag}. 
\end{itemize}

We present our results on zero-shot classification in Table~\ref{table:zero_shot_classification}. \HTLM{} prompting of classification datasets outperforms the most comparable (in terms of number of parameters) GPT-3 Medium sized model on the majority of tasks, while approaching---and on RTE outperforming---the GPT-3 Large model which consists of roughly double the amount of parameters as \HTLM{}.

\begin{table*}[htb]
\centering
\begin{tabular}{@{}lccccc@{}}
\toprule
 & \multicolumn{1}{l}{RTE} & \multicolumn{1}{l}{BoolQ} & \multicolumn{1}{l}{Winogrande} & HellaSwag & \# Params\\ \midrule
GPT-3        & 63.5 & 60.5 & 70.5 & 78.9 & 175B\\
GPT-3 Large  & 48.4 & 58.9 & 57.4 & 51.0 & 760M\\ 
GPT-3 Med  & 49.8 & 60.3 & 52.1 & 43.6 & 350M\\ \midrule
\HTLM{}-Manual & 51.2 & 55.3 & 54.8 & 47.9 & 400M\\ \bottomrule
\end{tabular}
\caption{Classification accuracy with zero shot prompting. We compare our performance to the full GPT-3 model as well as variants of comparable size.}
\label{table:zero_shot_classification}
\end{table*}

\section{Fine-tuning Experiments}
In addition to our previous prompting results, we also aim to show that \HTLM{} learned representations are useful in the full finetuning setting. We compare against other pre-training MLM models such as RoBERTa~\cite{ROBERTA}, original BART~\cite{BART}, and T5~\cite{T5} by finetuning on the GLUE benchmark~\citep{GLUE}. 

During finetuning, instead of a simple concatenation of sentences from the train set, we place the examples into prompts derived from \citet{how_many_datapoints}. We defer to the Appendix for the exact prompts. Given the recent advancements in finetuning, we also report results using the recently proposed R3F method for finetuning~\citep{RXF} for both RoBERTa and \HTLM{}.
\begin{table*}[ht]
\centering
\setlength{\tabcolsep}{3pt}
\begin{tabular}{@{}lcccccccc@{}}
\toprule
 &
  \begin{tabular}[c]{@{}l@{}}MNLI\\ Acc-m/mm\end{tabular} &
  \begin{tabular}[c]{@{}l@{}}QQP\\ Acc\end{tabular} &
  \begin{tabular}[c]{@{}l@{}}RTE\\ Acc\end{tabular} &
  \begin{tabular}[c]{@{}l@{}}QNLI\\ Acc\end{tabular} &
  \begin{tabular}[c]{@{}l@{}}MRPC\\ Acc\end{tabular} &
  \begin{tabular}[c]{@{}l@{}}CoLA\\ Mcc\end{tabular} &
  \begin{tabular}[c]{@{}l@{}}SST-2\\ Acc\end{tabular} & \quad \begin{tabular}[c]{@{}l@{}}\# Params\\ \quad\end{tabular}\\ \midrule

RoBERTA   & 90.2/-    & 92.2   & 86.6 & 94.7 & 89.1    & 68.0 & 96.4  & 330M  \\
RoBERTa-R3F        & 91.1/91.3         &  92.4         &  88.5    &     95.3  &   91.6        & \bf{71.2}      &  97.0    & 330M\\
T5-Base & 87.1/86.2 & 89.4 & 80.1 &  93.7 & 87.5 & 51.1 &  95.2 & 220M\\
T5-Large & 89.9/89.6 & 89.9 & 87.2 & 94.8 & 89.9 & 61.2 &  96.3 & 770M\\
BART-Large & 89.9/90.1 &  92.5 & 87.0 &  94.9 & 90.4 &  62.8 & 96.6 & 400M\\ \midrule
\HTLM{}        & 90.3/91.4  &   92.6       &  87.1    &   95.1   & 90.8    &  64.3    &   96.9 & 400M \\  \HTLM{}-R3F        & 91.4/92.1  &   92.8        &  89.1    &   95.4   & 91.5    &  69.4    &   97.1  & 400M \\
\HTLM{}-R3F-Prompt        & \bf{91.6/91.2}  &   \bf{92.9}        &  \bf{89.4}    &   \bf{95.7}   & \bf{91.7}    &  69.8    &   \bf{97.3}  & 400M\\ \bottomrule

\end{tabular}
\caption{Results on the GLUE development set for various fine-tuning methods
applied to \HTLM{}.}
\label{table:glue}
\end{table*}

We present our results in Table~\ref{table:glue}. Overall \HTLM{} improves over existing pre-training methods. We also note that we can improve fine-tuning performance by placing the examples into prompts and fine-tuning the classification head. The improvements that we see in terms of prompting have no adverse effects on fine-tuning but are rather positive, providing further evidence that the proposed approach of structured pre-training is a viable alternative to other methods of pre-training even for fine-tuning.

\ifshownlg We also show our fine-tuning results for the table-to-text generation datasets in Table~\ref{tab:table-to-text}. Similar to GLUE fine-tuning, we place all NLG samples into a prompt while fine-tuning. \HTLM{} fine-tuned is able to outperform both variants of the GPT-2 model consistently.\fi
\section{Prompt Data Efficiency}
\begin{table*}[htpb!]
\centering
\begin{tabular}{@{}l ccccc@{}}
\toprule 
& \multicolumn{5}{c}{Average Advantage (\# Training Points, P vs. H)} \\
& MNLI & BoolQ & CB & RTE & WiC\\\midrule 
RoBERTa-Large & $3506\pm536$ & $752\pm46$ & $90\pm2$ & $282\pm34$ & $-424\pm74$\\
T5-Large & $5010\pm230$ & $650\pm85$ & $150\pm8$ & $300\pm65$ & $-220\pm20$\\
BART-Large & $4020\pm220$ & $450\pm55$ & $125\pm10$ & $305\pm25$ & $-110\pm45$\\ \midrule
\HTLM{} & $\bf{6025\pm440}$ & $\bf{855\pm205}$ & $\bf{255\pm35}$ & $\bf{840\pm45}$ & $\bf{45\pm25}$\\
\bottomrule
\end{tabular}
\caption{Average advantage (higher is better)  in terms of training points for fine-tuning well-structured prompt ($P$) against a classical classification head ($H$).}
\label{table:data_points_per_prompt_p_h}
\end{table*}

\begin{table*}[htpb!]
\centering
\begin{tabular}{@{}l lcccc@{}}
\toprule 
& \multicolumn{5}{c}{Average Advantage (\# Training Points, P vs. N)} \\
& MNLI & BoolQ & CB & RTE & WiC\\\midrule 
RoBERTa-Large & $150\pm252$ & $299\pm81$ & $78\pm2$ & $404\pm68$ & $-354\pm166$\\
T5-Large & $300\pm120$ & $350\pm95$ & $150\pm4$ & $608\pm90$ & $20\pm43$\\
BART-Large & $200\pm180$ & $325\pm54$ & $85\pm8$ & $512\pm64$ & $-80\pm89$\\ \midrule
\HTLM{} & $\bf{692\pm240}$ & $\bf{565\pm143}$ & $\bf{255\pm34}$ & $\bf{640\pm45}$ & $\bf{80\pm40}$\\
\bottomrule
\end{tabular}
\caption{Average advantage (higher is better) in terms of training points for fine-tuning well-structured prompt ($P$) against a prompt with a non-sensical verbalizer ($N$). }
\label{table:data_points_per_prompt_p_n}
\end{table*}

What does the HTML-based pretraining and prompting scheme offer over one based on the plain text? \citet{how_many_datapoints} explored quantifying how many data points a single prompt was worth. Specifically, they analyzed three different task-specific settings given a pattern (the structure that the inputs are put into) and verbalizer (i.e., yes/no answer to pattern): (1) fine-tuning a classification head ($H$), (2) fine-tuning the verbalizer of a prompt encoding the semantics of the task ($P$), and (3) fine-tuning the prompt but with a verbalizer that is non-sensical ($N$).

By carefully selecting the number of data points to be used during training in each setting while matching the end fine-tuning performance, we can empirically measure the efficacy of prompts in terms of data points. We provide the same analysis extended to BART, T5-Large, and \HTLM{} using the same PET prompts provided in \citet{itsnotjustsize}. For \HTLM{}, we wrap all PET prompts in an HTML element. We select the same datasets that were used in the original paper for our experimentation; MNLI \citep{mnli}, BoolQ \citep{clark2019boolq}, CB \cite{demarneffe:cb}, RTE \citep{rte}, WiC \cite{pilehvar2018wic}.

We first look at the average advantage of fine-tuning a prompt ($P$) against a classification head ($H$) in Table~\ref{table:data_points_per_prompt_p_h}. We see that across the board, \HTLM{} prompts---i.e., hypertext prompts applied to \HTLM{}---are worth more than natural language prompts to various other pre-trained models. Compared to RoBERTa-Large on smaller datasets, \HTLM{}'s advantage is close to triple on CB and double on RTE. Furthermore, on WiC, \HTLM{} is the only pre-trained model capable of having a positive training advantage when using prompts. We view this as additional evidence to the benefit of pre-training on structured data on the prompting of a pre-trained model.

We also compare the average advantage of fine-tuning a prompt with a verbalizer ($P$) that makes sense against against finetuning a prompt where we change the verbalizer to a random first name ($N$). This is important to capture whether the benefits arise from representing the data in their respective patterns or the coupling of the pattern and the verbalizer. We present our results in Table~\ref{table:data_points_per_prompt_p_n}. Relative to the previous $P$ vs. $H$ setting we lose a large amount of advantage, as was similarly seen in \citep{how_many_datapoints}. Interestingly enough for small datasets such as CB, all of the training advantage of the prompt comes from the pattern in \HTLM{}.

We view this as further evidence that a structured, document level approach to both pre-training and prompting can be seen as a viable alternative to a purely natural language approach.

\section{Related Work}
GPT-2~\cite{GPT2} showed that large language models show varying levels of zero-shot performance across NLP tasks when compared to supervised baselines (e.g., rudimentary performance on summarization, but more competitive results on reading comprehension). 
\citet{gpt3} through their GPT3 model showed that by further scaling up language models on a large subset of the internet, prompting could be a viable alternative to standard finetuning. The success of GPT3 was largely attributed to massive size and compute-intensive pretraining. By reformulating NLP tasks as cloze-style questions, \citet{itsnotjustsize}  shows that the prompting capabilities exhibited by GPT3 can occur in language models of a much smaller scale when gradient-based finetuning is combined with task-specific unlabeled data. Follow-up work~\cite{adapet} improves upon these results without depending on unlabeled data. Unlike GPT-3 and other models which use conventional natural language text-based prompting, we focus on a new hyper-text based prompting scheme using generative masked language models pre-trained directly over HTML. 

For task-specific zero-shot performance, custom pre-training and data augmentation schemes have been developed. For example, PEGASUS~\citep{pegasus} proposes a novel pre-training scheme tailored for summarization which involves masking and generating salient \emph{gap} sentences from a large news corpus. While PEGASUS is capable of doing zero-shot summarization, it offers little control over summary attributes such as length and style which vary across different summarization datasets. 
WikiTransfer~\cite{Fabbri2021ImprovingZA}  fine-tunes pretrained models on pseudo-summaries, produced from generic Wikipedia data, which contain characteristics of the target dataset, such as the length and level of abstraction.
Our proposed model allows fine-grained control over the length of the generated text by specifying the size of the mask. Moreover, by using different prompts, \HTLM~can produce stylistically varied summaries without dataset-specific augmentation and finetuning. 

Another line of work has been looking at a hybrid form of prompting that attempts to optimize very few parameters to solve an end task. For example \citet{prefixtuning} argue that optimizing in the continuous prompt space is an effective solution to prompt search while \citet{intrinsic_dimensionality_finetuning} optimize for a low-rank projection of the full parameter space. For simplicity, we only focus on either full-finetuning or zero-shot prompting in this paper.

Attempts have been made to encode architectural priors for structured inputs into transformers as well. Specifically, \citet{etc_google} discuss a new type of model which allows for scalability in input length as well as the ability to encode the structure of the input. We opt to allow \HTLM{} to learn the structure that is available in the HTML directly without encoding any structural priors into the model itself.

\section{Conclusion}
In this paper, we proposed \HTLM{}, a hyper-text language model trained on simplified HTML documents from a large-scale web crawl. We showed that by directly modeling HTML through a BART-like objective, we could do structured zero-shot prompting by representing tasks in HTML. Specifically, we outperform the previous best results on zero-shot prompting for summarization by a wide margin by creating prompts that capture the underlying semantics of each summarization dataset. Furthermore, we show that pre-training on structured data improved full finetuning performance relative to other pre-trained models that only modeled natural language.

We also showed additional advantages of modeling hyper-text, beyond improved accuracy.  \HTLM{} can be used for auto-prompt  by simply asking the model to recover the document structure from training samples; these auto-prompts on datasets like Gigaword and CNN/DM outperformed previous state-of-the-art zero-shot approaches.
Lastly, we provided an in-depth comparison of the training advantage, in terms of data efficiency, that \HTLM{} had compared to other pre-training approaches. Across the board, HTML prompts were worth more to \HTLM{} than natural language prompts were worth to our baselines, further showing the efficacy of pre-training structured data.

Future work can focus on the scaling laws of structured pre-training and prompting. As was seen from GPT-3, the size of the model and the amount of compute utilized and significant impact on prompting performance.

\bibliography{anthology,acl2020}

\begin{thebibliography}{43}
\expandafter\ifx\csname natexlab\endcsname\relax\def\natexlab#1{#1}\fi

\bibitem[{Aghajanyan et~al.(2020{\natexlab{a}})Aghajanyan, Shrivastava, Gupta,
  Goyal, Zettlemoyer, and Gupta}]{RXF}
Armen Aghajanyan, Akshat Shrivastava, Anchit Gupta, Naman Goyal, Luke
  Zettlemoyer, and Sonal Gupta. 2020{\natexlab{a}}.
\newblock Better fine-tuning by reducing representational collapse.
\newblock \emph{arXiv preprint arXiv:2008.03156}.

\bibitem[{Aghajanyan et~al.(2020{\natexlab{b}})Aghajanyan, Zettlemoyer, and
  Gupta}]{intrinsic_dimensionality_finetuning}
Armen Aghajanyan, Luke Zettlemoyer, and Sonal Gupta. 2020{\natexlab{b}}.
\newblock Intrinsic dimensionality explains the effectiveness of language model
  fine-tuning.
\newblock \emph{arXiv preprint arXiv:2012.13255}.

\bibitem[{Ainslie et~al.(2020)Ainslie, Ontanon, Alberti, Cvicek, Fisher, Pham,
  Ravula, Sanghai, Wang, and Yang}]{etc_google}
Joshua Ainslie, Santiago Ontanon, Chris Alberti, Vaclav Cvicek, Zachary Fisher,
  Philip Pham, Anirudh Ravula, Sumit Sanghai, Qifan Wang, and Li~Yang. 2020.
\newblock Etc: Encoding long and structured inputs in transformers.
\newblock In \emph{Proceedings of the 2020 Conference on Empirical Methods in
  Natural Language Processing (EMNLP)}, pages 268--284.

\bibitem[{Beltagy et~al.(2020)Beltagy, Peters, and Cohan}]{longformer}
Iz~Beltagy, Matthew~E Peters, and Arman Cohan. 2020.
\newblock Longformer: The long-document transformer.
\newblock \emph{arXiv preprint arXiv:2004.05150}.

\bibitem[{Belz and Reiter(2006)}]{NIST}
Anja Belz and Ehud Reiter. 2006.
\newblock Comparing automatic and human evaluation of nlg systems.
\newblock In \emph{11th conference of the european chapter of the association
  for computational linguistics}.

\bibitem[{Bentivogli et~al.(2009)Bentivogli, Clark, Dagan, and
  Giampiccolo}]{rte}
Luisa Bentivogli, Peter Clark, Ido Dagan, and Danilo Giampiccolo. 2009.
\newblock The fifth pascal recognizing textual entailment challenge.
\newblock In \emph{TAC}.

\bibitem[{Brown et~al.(2020)Brown, Mann, Ryder, Subbiah, Kaplan, Dhariwal,
  Neelakantan, Shyam, Sastry, Askell et~al.}]{gpt3}
Tom~B Brown, Benjamin Mann, Nick Ryder, Melanie Subbiah, Jared Kaplan, Prafulla
  Dhariwal, Arvind Neelakantan, Pranav Shyam, Girish Sastry, Amanda Askell,
  et~al. 2020.
\newblock Language models are few-shot learners.
\newblock \emph{arXiv preprint arXiv:2005.14165}.

\bibitem[{Choromanski et~al.(2020)Choromanski, Likhosherstov, Dohan, Song,
  Gane, Sarlos, Hawkins, Davis, Mohiuddin, Kaiser et~al.}]{performer}
Krzysztof Choromanski, Valerii Likhosherstov, David Dohan, Xingyou Song,
  Andreea Gane, Tamas Sarlos, Peter Hawkins, Jared Davis, Afroz Mohiuddin,
  Lukasz Kaiser, et~al. 2020.
\newblock Rethinking attention with performers.
\newblock \emph{arXiv preprint arXiv:2009.14794}.

\bibitem[{Clark et~al.(2019)Clark, Lee, Chang, Kwiatkowski, Collins, and
  Toutanova}]{clark2019boolq}
Christopher Clark, Kenton Lee, Ming-Wei Chang, Tom Kwiatkowski, Michael
  Collins, and Kristina Toutanova. 2019.
\newblock {B}ool{Q}: Exploring the surprising difficulty of natural yes/no
  questions.
\newblock In \emph{Proceedings of NAACL-HLT 2019}.

\bibitem[{Conneau et~al.(2019)Conneau, Khandelwal, Goyal, Chaudhary, Wenzek,
  Guzm{\'a}n, Grave, Ott, Zettlemoyer, and Stoyanov}]{XLMR}
Alexis Conneau, Kartikay Khandelwal, Naman Goyal, Vishrav Chaudhary, Guillaume
  Wenzek, Francisco Guzm{\'a}n, Edouard Grave, Myle Ott, Luke Zettlemoyer, and
  Veselin Stoyanov. 2019.
\newblock Unsupervised cross-lingual representation learning at scale.
\newblock \emph{arXiv preprint arXiv:1911.02116}.

\bibitem[{De~Marneffe et~al.(2019)De~Marneffe, Simons, and
  Tonhauser}]{demarneffe:cb}
Marie-Catherine De~Marneffe, Mandy Simons, and Judith Tonhauser. 2019.
\newblock {The CommitmentBank}: Investigating projection in naturally occurring
  discourse.
\newblock To appear in proceedings of Sinn und Bedeutung 23. Data can be found
  at https://github.com/mcdm/CommitmentBank/.

\bibitem[{Fabbri et~al.(2021)Fabbri, Han, Li, Li, Ghazvininejad, Joty, Radev,
  and Mehdad}]{Fabbri2021ImprovingZA}
A.~R. Fabbri, Simeng Han, Haoyuan Li, Haoran Li, Marjan Ghazvininejad,
  Shafiq~R. Joty, Dragomir Radev, and Yashar Mehdad. 2021.
\newblock Improving zero and few-shot abstractive summarization with
  intermediate fine-tuning and data augmentation.
\newblock In \emph{NAACL}.

\bibitem[{Gardent et~al.(2017)Gardent, Shimorina, Narayan, and
  Perez-Beltrachini}]{webnlg}
Claire Gardent, Anastasia Shimorina, Shashi Narayan, and Laura
  Perez-Beltrachini. 2017.
\newblock The webnlg challenge: Generating text from rdf data.
\newblock In \emph{Proceedings of the 10th International Conference on Natural
  Language Generation}, pages 124--133.

\bibitem[{Hermann et~al.(2015)Hermann, Kocisky, Grefenstette, Espeholt, Kay,
  Suleyman, and Blunsom}]{cnndailymail}
Karl~Moritz Hermann, Tomas Kocisky, Edward Grefenstette, Lasse Espeholt, Will
  Kay, Mustafa Suleyman, and Phil Blunsom. 2015.
\newblock Teaching machines to read and comprehend.
\newblock In \emph{Advances in neural information processing systems}, pages
  1693--1701.

\bibitem[{Joulin et~al.(2016)Joulin, Grave, Bojanowski, Douze, J{\'e}gou, and
  Mikolov}]{fasttext}
Armand Joulin, Edouard Grave, Piotr Bojanowski, Matthijs Douze, H{\'e}rve
  J{\'e}gou, and Tomas Mikolov. 2016.
\newblock Fasttext. zip: Compressing text classification models.
\newblock \emph{arXiv preprint arXiv:1612.03651}.

\bibitem[{Kim et~al.(2018)Kim, Kim, and Kim}]{reddittifu}
Byeongchang Kim, Hyunwoo Kim, and Gunhee Kim. 2018.
\newblock Abstractive summarization of reddit posts with multi-level memory
  networks.
\newblock \emph{arXiv preprint arXiv:1811.00783}.

\bibitem[{Kingma and Ba(2014)}]{ADAM}
Diederik~P Kingma and Jimmy Ba. 2014.
\newblock Adam: A method for stochastic optimization.
\newblock \emph{arXiv preprint arXiv:1412.6980}.

\bibitem[{Lavie and Agarwal(2007)}]{METEOR}
Alon Lavie and Abhaya Agarwal. 2007.
\newblock Meteor: An automatic metric for mt evaluation with high levels of
  correlation with human judgments.
\newblock In \emph{Proceedings of the second workshop on statistical machine
  translation}, pages 228--231.

\bibitem[{Le~Scao and Rush(2021)}]{how_many_datapoints}
Teven Le~Scao and Alexander Rush. 2021.
\newblock \href {https://doi.org/10.18653/v1/2021.naacl-main.208} {How many
  data points is a prompt worth?}
\newblock In \emph{Proceedings of the 2021 Conference of the North American
  Chapter of the Association for Computational Linguistics: Human Language
  Technologies}, pages 2627--2636, Online. Association for Computational
  Linguistics.

\bibitem[{Levesque et~al.(2012)Levesque, Davis, and Morgenstern}]{winograd}
Hector Levesque, Ernest Davis, and Leora Morgenstern. 2012.
\newblock The winograd schema challenge.
\newblock In \emph{Thirteenth International Conference on the Principles of
  Knowledge Representation and Reasoning}. Citeseer.

\bibitem[{Levesque et~al.(2011)Levesque, Davis, and
  Morgenstern}]{levesque2011winograd}
Hector~J Levesque, Ernest Davis, and Leora Morgenstern. 2011.
\newblock The {W}inograd schema challenge.
\newblock In \emph{{AAAI} Spring Symposium: Logical Formalizations of
  Commonsense Reasoning}, volume~46, page~47.

\bibitem[{Lewis et~al.(2019)Lewis, Liu, Goyal, Ghazvininejad, Mohamed, Levy,
  Stoyanov, and Zettlemoyer}]{BART}
Mike Lewis, Yinhan Liu, Naman Goyal, Marjan Ghazvininejad, Abdelrahman Mohamed,
  Omer Levy, Ves Stoyanov, and Luke Zettlemoyer. 2019.
\newblock Bart: Denoising sequence-to-sequence pre-training for natural
  language generation, translation, and comprehension.
\newblock \emph{arXiv preprint arXiv:1910.13461}.

\bibitem[{Li and Liang(2021)}]{prefixtuning}
Xiang~Lisa Li and Percy Liang. 2021.
\newblock Prefix-tuning: Optimizing continuous prompts for generation.
\newblock \emph{arXiv preprint arXiv:2101.00190}.

\bibitem[{Lin(2004)}]{ROUGE}
Chin-Yew Lin. 2004.
\newblock Rouge: A package for automatic evaluation of summaries.
\newblock In \emph{Text summarization branches out}, pages 74--81.

\bibitem[{Liu et~al.(2019)Liu, Ott, Goyal, Du, Joshi, Chen, Levy, Lewis,
  Zettlemoyer, and Stoyanov}]{ROBERTA}
Yinhan Liu, Myle Ott, Naman Goyal, Jingfei Du, Mandar Joshi, Danqi Chen, Omer
  Levy, Mike Lewis, Luke Zettlemoyer, and Veselin Stoyanov. 2019.
\newblock Roberta: A robustly optimized bert pretraining approach.
\newblock \emph{arXiv preprint arXiv:1907.11692}.

\bibitem[{Nan et~al.(2020)Nan, Radev, Zhang, Rau, Sivaprasad, Hsieh, Tang,
  Vyas, Verma, Krishna et~al.}]{dartnlg}
Linyong Nan, Dragomir Radev, Rui Zhang, Amrit Rau, Abhinand Sivaprasad,
  Chiachun Hsieh, Xiangru Tang, Aadit Vyas, Neha Verma, Pranav Krishna, et~al.
  2020.
\newblock Dart: Open-domain structured data record to text generation.
\newblock \emph{arXiv preprint arXiv:2007.02871}.

\bibitem[{Napoles et~al.(2012)Napoles, Gormley, and Van~Durme}]{gigaword}
Courtney Napoles, Matthew~R Gormley, and Benjamin Van~Durme. 2012.
\newblock Annotated gigaword.
\newblock In \emph{Proceedings of the Joint Workshop on Automatic Knowledge
  Base Construction and Web-scale Knowledge Extraction (AKBC-WEKEX)}, pages
  95--100.

\bibitem[{Narayan et~al.(2018)Narayan, Cohen, and Lapata}]{xsum}
Shashi Narayan, Shay~B Cohen, and Mirella Lapata. 2018.
\newblock Don't give me the details, just the summary! topic-aware
  convolutional neural networks for extreme summarization.
\newblock \emph{arXiv preprint arXiv:1808.08745}.

\bibitem[{Novikova et~al.(2017)Novikova, Du{\v{s}}ek, and Rieser}]{e2e_nlg}
Jekaterina Novikova, Ond{\v{r}}ej Du{\v{s}}ek, and Verena Rieser. 2017.
\newblock The e2e dataset: New challenges for end-to-end generation.
\newblock \emph{arXiv preprint arXiv:1706.09254}.

\bibitem[{Papineni et~al.(2002)Papineni, Roukos, Ward, and Zhu}]{bleu}
Kishore Papineni, Salim Roukos, Todd Ward, and Wei-Jing Zhu. 2002.
\newblock Bleu: a method for automatic evaluation of machine translation.
\newblock In \emph{Proceedings of the 40th annual meeting of the Association
  for Computational Linguistics}, pages 311--318.

\bibitem[{Perez et~al.(2021)Perez, Kiela, and Cho}]{true_few_shot_learning}
Ethan Perez, Douwe Kiela, and Kyunghyun Cho. 2021.
\newblock True few-shot learning with language models.
\newblock \emph{arXiv preprint arXiv:2105.11447}.

\bibitem[{Pilehvar and Camacho-Collados(2019)}]{pilehvar2018wic}
Mohammad~Taher Pilehvar and Jose Camacho-Collados. 2019.
\newblock {WiC}: The word-in-context dataset for evaluating context-sensitive
  meaning representations.
\newblock In \emph{Proceedings of NAACL-HLT}.

\bibitem[{Radford et~al.(2019)Radford, Wu, Child, Luan, Amodei, and
  Sutskever}]{GPT2}
Alec Radford, Jeffrey Wu, Rewon Child, David Luan, Dario Amodei, and Ilya
  Sutskever. 2019.
\newblock Language models are unsupervised multitask learners.
\newblock \emph{OpenAI Blog}, 1(8):9.

\bibitem[{Raffel et~al.(2019)Raffel, Shazeer, Roberts, Lee, Narang, Matena,
  Zhou, Li, and Liu}]{T5}
Colin Raffel, Noam Shazeer, Adam Roberts, Katherine Lee, Sharan Narang, Michael
  Matena, Yanqi Zhou, Wei Li, and Peter~J Liu. 2019.
\newblock Exploring the limits of transfer learning with a unified text-to-text
  transformer.
\newblock \emph{arXiv preprint arXiv:1910.10683}.

\bibitem[{Schick and Sch{\"u}tze(2020)}]{itsnotjustsize}
Timo Schick and Hinrich Sch{\"u}tze. 2020.
\newblock It's not just size that matters: Small language models are also
  few-shot learners.
\newblock \emph{arXiv preprint arXiv:2009.07118}.

\bibitem[{Snover et~al.(2005)Snover, Dorr, Schwartz, Makhoul, Micciulla, and
  Weischedel}]{ter}
Mathew Snover, Bonnie Dorr, Richard Schwartz, John Makhoul, Linnea Micciulla,
  and Ralph Weischedel. 2005.
\newblock A study of translation error rate with targeted human annotation.
\newblock In \emph{Proceedings of the 7th Conference of the Association for
  Machine Translation in the Americas (AMTA 06)}, pages 223--231.

\bibitem[{Tam et~al.(2021)Tam, Menon, Bansal, Srivastava, and Raffel}]{adapet}
Derek Tam, R.~R. Menon, M.~Bansal, Shashank Srivastava, and Colin Raffel. 2021.
\newblock Improving and simplifying pattern exploiting training.
\newblock \emph{ArXiv}, abs/2103.11955.

\bibitem[{Vedantam et~al.(2015)Vedantam, Lawrence~Zitnick, and Parikh}]{cider}
Ramakrishna Vedantam, C~Lawrence~Zitnick, and Devi Parikh. 2015.
\newblock Cider: Consensus-based image description evaluation.
\newblock In \emph{Proceedings of the IEEE conference on computer vision and
  pattern recognition}, pages 4566--4575.

\bibitem[{Wang et~al.(2018)Wang, Singh, Michael, Hill, Levy, and Bowman}]{GLUE}
Alex Wang, Amanpreet Singh, Julian Michael, Felix Hill, Omer Levy, and Samuel
  Bowman. 2018.
\newblock \href {https://doi.org/10.18653/v1/W18-5446} {{GLUE}: A multi-task
  benchmark and analysis platform for natural language understanding}.
\newblock In \emph{Proceedings of the 2018 {EMNLP} Workshop {B}lackbox{NLP}:
  Analyzing and Interpreting Neural Networks for {NLP}}, pages 353--355,
  Brussels, Belgium. Association for Computational Linguistics.

\bibitem[{Wang et~al.(2020)Wang, Li, Khabsa, Fang, and Ma}]{linformer}
Sinong Wang, Belinda Li, Madian Khabsa, Han Fang, and Hao Ma. 2020.
\newblock Linformer: Self-attention with linear complexity.
\newblock \emph{arXiv preprint arXiv:2006.04768}.

\bibitem[{Williams et~al.(2018)Williams, Nangia, and Bowman}]{mnli}
Adina Williams, Nikita Nangia, and Samuel Bowman. 2018.
\newblock \href {http://aclweb.org/anthology/N18-1101} {A broad-coverage
  challenge corpus for sentence understanding through inference}.
\newblock In \emph{Proceedings of the 2018 Conference of the North American
  Chapter of the Association for Computational Linguistics: Human Language
  Technologies, Volume 1 (Long Papers)}, pages 1112--1122. Association for
  Computational Linguistics.

\bibitem[{Zellers et~al.(2019)Zellers, Holtzman, Bisk, Farhadi, and
  Choi}]{hellaswag}
Rowan Zellers, Ari Holtzman, Yonatan Bisk, Ali Farhadi, and Yejin Choi. 2019.
\newblock Hellaswag: Can a machine really finish your sentence?
\newblock \emph{arXiv preprint arXiv:1905.07830}.

\bibitem[{Zhang et~al.(2019)Zhang, Zhao, Saleh, and Liu}]{pegasus}
Jingqing Zhang, Yao Zhao, Mohammad Saleh, and Peter~J Liu. 2019.
\newblock Pegasus: Pre-training with extracted gap-sentences for abstractive
  summarization.
\newblock \emph{arXiv preprint arXiv:1912.08777}.

\end{thebibliography}
\bibliographystyle{acl_natbib}

\clearpage
\appendix
\section{Appendix}
\subsection{Finetuning Hyper-Parameters}
\begin{table*}[!htpb]
\centering
\begin{tabular}{@{}llllllll@{}}
\toprule
Hyper Parameter & MNLI   & QNLI  & QQP    & SST-2 & RTE  & MRPC & CoLA \\ \midrule
Learning Rate   & 5e-6   & 5e-6  & 5e-6   & 5e-6  & 1e-5 & 1e-5 & 1e-5 \\
Max Updates     & 123873 & 33112 & 113272 & 20935 & 3120 & 2296 & 5336 \\
Max Sentences   & 8      & 8     & 32     & 32    & 8    & 16   & 16   \\
\bottomrule  
\end{tabular}
\caption{Task specific hyper parameters for GLUE experiments}
\end{table*}

\begin{table*}[!htpb]
\centering
\begin{tabular}{@{}ll@{}}
\toprule
Hyper parameter & Value              \\ \midrule
Optimizer       & Adam               \\
Adam-betas      & (0.9, 0.98)        \\
Adam-eps        & 1e-6               \\
LR Scheduler    & polynomial decay   \\
Dropout         & 0.1                \\
Weight Decay    & 0.01               \\
Warmup Updates  & 0.06 * max updates \\
\bottomrule
\end{tabular}
\quad
\begin{tabular}{@{}ll@{}}
\toprule
Hyper parameter & Value           \\ \midrule
$\lambda$          & [0.1, 0.5, 1.0, 5.0] \\
Noise Types     & [$\mathcal{U}$, $\mathcal{N}$] \\
$\sigma$        & $1e-5$\\ 
\bottomrule
\end{tabular}
\caption{Hyper parameters for R3F experiments on GLUE}
\end{table*}
For our GLUE related experiments the following parameters are used.

\end{document}